\documentclass[acmtog]{acmart}
\AtBeginDocument{%
  \providecommand\BibTeX{{%
    \normalfont B\kern-0.5em{\scshape i\kern-0.25em b}\kern-0.8em\TeX}}}

\usepackage{stackengine}
\usepackage{caption}
\usepackage{subcaption}
\usepackage{algorithm}
\usepackage{algpseudocode}
\usepackage[utf8]{inputenc} 
\usepackage[T1]{fontenc}    
\usepackage{hyperref}       
\usepackage{url}            
\usepackage{booktabs}       
\usepackage{amsfonts}       
\usepackage{nicefrac}       
\usepackage{microtype}      
\usepackage{xcolor}         
\usepackage{multirow}
\usepackage{svg}
\usepackage{amsmath}

\usepackage{longtable}
\usepackage{lscape}
\usepackage{cleveref}
\usepackage{adjustbox}
\usepackage{csquotes}
\usepackage{graphicx,wrapfig,lipsum}

\usepackage{dsfont}
\newcommand{\indicator}{\mathds{1}}

\makeatletter
\newcommand{\printfnsymbol}[1]{%
  \textsuperscript{\@fnsymbol{#1}}%
}
\makeatother

\setcopyright{acmcopyright}
\copyrightyear{2022}
\acmYear{2022}

\begin{document}

\title{Fairness Implications of Encoding Protected Categorical Attributes}

\author{Carlos Mougan}
\affiliation{%
  \institution{University of Southampton}
  \country{United Kingdom}}
\email{c.mougan-navarro@soton.ac.uk}

\author{Jose M. Alvarez}
\affiliation{%
  \institution{Scuola Normale Superiore}
  \country{Italy}
}
\affiliation{%
  \institution{University of Pisa}
  \country{Italy}
}
\email{jose.alvarez@sns.it}


\author{Salvatore Ruggieri}
\affiliation{%
  \institution{University of Pisa}
  \country{Italy}
}
\email{salvatore.ruggieri@unipi.it}
\author{Steffen Staab}
\affiliation{%
  \institution{University of Southampton}
  \country{United Kingdom}
}
\affiliation{%
  \institution{University of Stuttgart}
  \country{Germany}
}
\email{S.R.Staab@soton.ac.uk }

\renewcommand{\shortauthors}{Mougan C., Alvarez J.,  Ruggieri S. and Staab S.}

\definecolor{amethyst}{rgb}{0.6, 0.4, 0.8}
\newcommand{\gourab}[1]{\textcolor{blue}{GP:#1}}
\newcommand{\steffen}[1]{\textcolor{red}{Steffen:#1}}
\newcommand{\carlos}[1]{\textcolor{amethyst}{Carlos:#1}}
\newcommand{\sr}[1]{\textcolor{red}{Salvatore:#1}}
\newcommand{\jose}[1]{\textcolor{green}{Jose:#1}}

\begin{abstract}
Past research has demonstrated that the explicit use of protected attributes in machine learning can improve both performance and fairness. Many machine learning algorithms, however, cannot directly process categorical attributes, such as country of birth or ethnicity. Because protected attributes frequently are categorical, they must be encoded  as features that can be input to  a chosen machine learning algorithm, e.g.\ support vector machines, gradient boosting decision trees or linear models. Thereby, encoding methods influence how and what  the machine learning algorithm will learn, affecting  model performance and fairness.This work compares the accuracy and fairness implications of the two most well-known encoding methods: \emph{one-hot encoding} and \emph{target encoding}. We distinguish between two types of induced bias that may arise from these encoding methods and may lead to unfair models. The first type, \textit{irreducible bias}, is due to direct group category discrimination and the second type, \textit{reducible bias}, is due to large variance in  statistically underrepresented groups. We investigate the interaction between categorical encodings and target encoding regularization methods that reduce  unfairness. Furthermore, we consider the problem of intersectional unfairness that may arise when machine learning best-practices improve performance measures by encoding several categorical attributes into a high-cardinality feature.




\end{abstract}

\ccsdesc[500]{Computing methodologies~Supervised learning by classification}
\ccsdesc[500]{Computing methodologies~Classification and regression trees}
\ccsdesc[500]{Social and professional topics~Socio-technical systems}

\ccsdesc[500]{Computing methodologies~Supervised learning by regression}

\keywords{Fairness, Algorithmic Accountability, Categorical Features, Bias}
\maketitle

\section{Introduction}
Anti-discrimination laws~\cite{gdpr,EU_fundamentalrights,Barocas2016BigDataImpact} prohibit the unfair treatment of individuals based on \emph{sensitive attributes} (also referred to as \emph{protected attributes}). The list of sensitive attributes varies per country, though these usually include gender, ethnicity, and religion~\cite{RomeiR2014_DiscSurvey}.
Following such legal motivations along with societal expectations, many studies have looked into discrimination in machine learning and proposed various ways to promote fairness (e.g.,~\cite{DBLP:journals/csur/MehrabiMSLG21,finocchiaro2021bridging, Kdd_Pedreschi2008_DataMiningDiscrimination, Ruggieri2010_DMforDD}).

The handling of sensitive attributes throughout the machine learning pipeline is central to establishing fairness. An early common practice was removing data on sensitive attributes altogether. This technique has been questioned because sensitive  attributes may be required for avoiding discrimination in data-driven decision models~\citep{DBLP:journals/ail/ZliobaiteC16,DBLP:journals/kais/KamiranC11}. Therefore, later work~\cite{zafar2017fairness,hartdt_equality,zafar2017parity,DBLP:conf/fat/MenonW18} has aimed at how to obtain fairer models given the presence of sensitive attributes, formalizing the problem as an optimization trade-off between model quality in terms of performance and some fairness objective.

Sensitive attributes often come as categorical data. For instance, roughly $75\%$ of the famous COMPAS dataset~\cite{compass} consists of categorical attributes, including most of the sensitive ones (see Section~\ref{sec:experiments.compass} for more details). Many machine learning algorithms require categorical attributes to be suitably \emph{encoded} as numerical data. Different ways of encoding categorical attributes into numerical features~\cite{high-cardinality-categorical,quantile2021,statisticallearning} have been proposed and extensively studied in the literature along with statistical regularization methods since the mid-1950s~\cite{pargent2021regularized}. This has resulted in various  methods  that encode categorical attributes as numerical data to make them usable by popular machine learning models, such as support vector machines, gradient boosting decision trees or linear models.

In this paper, \textit{we study the broader implications that encoding categorical sensitive attributes can have on model accuracy and fairness}. Despite being a common machine learning practice, often falling within the data pre-processing step, the effects of categorical attribute encodings on fairness remain largely unexplored. For instance, prior works on fair machine learning~\cite{zafar2017fairness,hartdt_equality,zafar2017parity} also use encoding of protected categorical attributes without any discussion on its implications or on the choice of encoding.

We focus on the two most widely used encodings: \textit{one-hot encoding} and \textit{target encoding}~\cite{quantile2021,category_encoders,catboost-encoder,pargent2021regularized}. \emph{One-hot encoding}, an unsupervised technique, produces orthogonal and equidistant vectors for each category~\cite{natureOHE,quantile2021}, thereby considering the categories to be equally independent of each other and other attributes.
However, when dealing with high cardinality categorical variables, one-hot encoding suffers from a lack of scalability and sparsity issues  due to the creation of many orthogonal dimensions (later discussed in Section~\ref{sec:relatedwork} and ~\ref{sec:methodology}). \emph{Target encoding}~\cite{quantile2021,high-cardinality-categorical,pargent2021regularized} is a supervised technique 
that replaces a categorical attribute with the mean target value of each corresponding category. Thus, it can handle all categories together in one dimension.\footnote{Target encoding methods have become an industry standard for high-cardinality categorical data~\citep{quantile2021,pargent2021regularized,target_embedding,howtowinKaggle} with algorithmic procedures being implemented in many open source packages~\cite{sktools,category_encoders}. One of the most common is the Python package called  \textit{category encoders} (\url{https://pepy.tech/project/category_encoders}). It achieves up to 1 million downloads per month. Target encoding is  the default encoding method in some high-performance open-source software implementations such as \textit{catboost}~\cite{hancock2020catboost,dorogush2018catboost} that has reached a total download of 74 millions.} 


The first problem of unfairness related to sensitive categorical attributes, which we call \emph{irreducible bias}, is related to the statistical differences between two highly populated groups: more data about the compared groups will not diminish this type of bias.  The second problem  arises because sampling from small groups may exhibit large variance leading to unfairness constituting  \emph{ reducible bias}. Many datasets contain  distributions of data that are imbalanced over different values of categorical attributes, often leading to performance degradation of the learned models (known as the {\it classes imbalance problem}~\cite{japkowicz2002class}).
The use of encodings in such datasets may naturally introduce disparity as per the observed class imbalance.
%
Moreover, when a dataset contains several sensitive categorical attributes, and these are merged to become one feature (a strategy often followed to improve model quality substantially~\cite{howtowinKaggle}), encodings may create fine-grained, sparsely populated intersectional features~\cite{kearns2018preventing,kearns2018empirical,DBLP:conf/fat/WangRR22}  increasing the chance for both types of induced biases~\cite{DBLP:conf/sdm/FouldsIKP20}.

%

The effects of encodings on model quality and fairness under the interplay of different encoding and regularization techniques have not been studied in the literature, although they affect very commonly used machine learning practices.
For target encoding, we study two popular statistical regularization methods called \textit{smoothing} and \textit{Gaussian noise regularization}, 
which provides a new avenue for the analysis of implications of categorical encodings on fairness. Through both a theoretical analysis as well as an empirical analysis using two real-world datasets, we find that suitable regularization can address unfairness   arising from the target encodings with only marginal losses in accuracy.\footnote{In this work we use the term ``accuracy'' in a non-literal sense to refer to the model performance, rather than the statistical evaluation metric of the same name. In Section~\ref{sec:methodology.metrics}, we discuss the adequate evaluation metric for our scenarios.} 
In summary, we make the following contributions:
\begin{itemize}
    
    \item We compare the two best-known categorical feature encoding methods, one-hot encoding, and target encoding against learning without protected attribute(s), in terms of model performance and fairness.
    
    \item We study the relation between regularization of target encodings and fairness by evaluating smoothing and Gaussian noise, which are two common techniques used for regularization by data preprocessing. 

    \item We provide evidence that  creating intersectional features can worsen discrimination. We show that a regularized target encoder can retain the benefits of intersectional features without increasing unfair discrimination.
    
    \item We  provide a theoretical analysis studing two types of induced biases, irreducible and reducible, that arise while encoding categorical protected attributes. 
\end{itemize}

\section{Background}
\label{sec:relatedwork}

\subsection{Categorical attribute encoding}
Handling categorical attributes is a common problem in machine learning, given that many algorithms use numerical data~\cite{tutz_2011,quantile2021}. There are many well-known methods for approaching this problem~\cite{Pargent:2019,10.1007/978-3-031-19493-1_14,DBLP:journals/tkde/CerdaV22,DBLP:journals/ml/CerdaVK18,barragan2022tratamiento}. 


\emph{One-hot encoding} (also known as dummy variables in the social sciences~\citep{Wooldridge2015_IntroductoryMetrics}) constructs orthogonal and equidistant vectors for each category. Given high cardinality categorical attributes, one-hot encoding suffers from shortcomings: (i) the dimension of the input space increases with the cardinality of the encoded variable, (ii) the derived features are rarely non-zero, and new and unseen categories cannot be handled~\cite{catFeatBayesian,quantile2021}.

 \textit{Label/ordinal encoding}~\cite{ordinal} uses a range of integers to represent  different categorical values. These are assumed to have no true order and integers are assigned in the order of appearance of the categories. Label encoding suffers less from  higher cardinalities of attribute values, but imposes an artificial random order on the categories, which may harm  learning. This, in turn, obstructs the model to extract meaningful information from the categorical data.
\begin{table}[ht]
    \centering
    \caption{An illustrative example of one-hot and target encoding methods over the same data sample.}\label{tab:encoding}
    \begin{tabular}{c|c|c}
            \textbf{Ethnic}  & \textbf{Encoding} & \textbf{Label} \\ \hline
            African-American & 1                 & 1              \\
            Caucasian        & $1/3$             & 1\\
            Caucasian        & $1/3$             & 0\\
            Caucasian        & $1/3$             & 0 \\
            Hispanic         & 0                 & 0            
    \end{tabular}
    \caption*{$(a)$ Unregularized Mean Target Encoding}
\end{table}
\begin{table}[ht]
    \centering
    
            \begin{tabular}{c|ccc}
            \textbf{Ethnic}  & \textbf{African-} & \textbf{Caucasian} & \textbf{Hispanic} \\
            &\textbf{American}&&\\ \hline
            African-American & 1                                & 0                          & 0                        \\
            Caucasian        & 0                                & 1                          & 0                        \\
            Caucasian        & 0                                & 1                          & 0                        \\
            Caucasian        & 0                                & 1                          & 0                        \\
            Hispanic         & 0                                & 0                          & 1                       
            \end{tabular}
            \caption*{$(b)$ One Hot Encoding}
\end{table}

\emph{Target encoding} replaces  attribute categories by the mean target value\footnote{Throughout the paper, we assume a binary target feature with values $\{0, 1\}$.} of each corresponding category. Thus, the high cardinality problem is addressed and categories are ordered in a meaningful manner~\cite{high-cardinality-categorical,DBLP:journals/tkde/CerdaV22}. The main drawback of target encoding appears when
the  target values of a
category with few samples are averaged.
The model may overly rely on the resulting target value  potentially suffering from inherent variance in
the small sample of data points from this category.
To overcome this problem several strategies introduce regularization terms in the target estimation~\cite{high-cardinality-categorical,quantile2021,pargent2021regularized}.

In \Cref{tab:encoding}, we illustrate one-hot encoding and target encoding for the category ethnicity using a 5 person sample  from the COMPAS dataset~\cite{compass}. The problem of over-fitting is evident for the cases of \textit{African-American} and \textit{Hispanic} where their encoding is replaced directly with the target, creating a data leakage that can potentially cause reducible induced bias (cf.\ Section~\ref{sec:experiments}).

Even though, early works that have studied preprocessing techniques for classification without discrimination ~\cite{DBLP:journals/kais/KamiranC11}, do not discuss the fairness effects of encoding categorical protected attributes.To the best of our knowledge, there is no previous work that studies the different effects of regularization on target encodings, nor the fairness implications of encoding categorical protected attributes. 

\subsection{Group Fairness}
Various definitions of fairness in machine learning have been proposed (see, e.g.,~\cite{DBLP:journals/csur/MehrabiMSLG21,finocchiaro2021bridging, barocas-hardt-narayanan} for recent overviews).
They can be categorized into notions of individual fairness and group fairness.
While metrics of individual fairness judge whether {\it  similar individuals are treated similarly}~\cite{DworkHPRZ2012_IndFair}, metrics of group fairness measure disparate treatment of groups, which are assembled according to shared categories of sensitive attributes of their individuals such as gender or race~\cite{ruggieri23}.

Different disparity metrics emphasize varying aspects of disparate treatment. For comprehensive understanding,  we investigate the effects of encoding methods according to three common disparity metrics, i.e.\ equal opportunity, statistical disparity (demographic disparity) and average absolute odds (equalized odds). All three metrics indicate equal treatments of different groups by values close to zero and highly disparate treatments by values different from zero.


In this work, we distinguish and define two types of induced bias or discrimination that the encoding of categorical attributes introduces.
Borrowing terminology used with regard to different types of uncertainty~\cite{aleatory_epistemic,Gal2016Uncertainty,epistemic_uncertainty}, we use \textit{Irreducible bias}   to refer to (direct) group discrimination arising from the categorization of groups into labels: more data about the compared groups do not reduce this type of bias. \textit{Reducible} bias arises due when the variance of  categories  with few instances cannot be well contained.

\subsection{Addressing Intersectionality}
\label{sec:relatedwork.intersectional}


It is a common trick for boosting model performance to concatenate multiple categorical variables and encode them into a single feature~\cite{howtowinKaggle}. This feature engineering procedure, which includes target encoding, parallels a possible implementation of \textit{intersectionality} when we concatenated two or more protected attributes. Intersectionality refers to when an individual that belongs to more than one protected group experiences discrimination at the intersection of these groups. It broadly refers to how different identities interact to produce a unique new form of discrimination \cite{DBLP:conf/fat/WangRR22}. \citet{crenshaw2013demarginalizing}, for example, studied how \textit{black women} in the United States experience discrimination beyond being either black or women. 

Although individuals often belong to multiple protected groups, intersectionality is largely understudied within algorithmic fairness. With some exceptions (e.g.,~\cite{FouldsIKP20_IntersectionalDefFairness, YangLS21_CausalIntersectionality,DBLP:conf/fat/WangRR22,DBLP:conf/cvpr/YucerAMB20, Alvarez2023_CST}), most works assume the single binary protected attribute or disregard intersectionality entirely when handling multiple protected attributes \cite{DBLP:conf/fat/WangRR22}, which is unrealistic and reductive. 
This is a pertinent issue as it is possible for individuals not to suffer from multiple discrimination but to suffer from intersectional discrimination \cite{Xenidis2020_TunningEULaw, DBLP:journals/corr/abs-2302-05995}.

We address the intersectionality concerns linked to target encoding. One hand, it can boost model performance; on the other hand, it can introduce new forms of discrimination. We add to this small but growing fairness literature by analyzing how target encoding can enable an implementation of intersectionality. In particular, we study how target encoding regularization can mitigate the potential biases induced by this feature engineering practice as well as compare it to the standard alternative of one-hot encoding.

\section{Formalization and Regularization of target encoding}
\label{sec:methodology}\label{sec:regularization}

Consider a categorical attribute $Z$ with domain $dom(Z) = \{z_1, \ldots,$ $z_c\}$, a binary target attribute $Y$ with $dom(Y) = \{0, 1\}$, and the joint probability of $P(Z, Y)$ over the population of interest. Target encoding replaces $Z$ with a continuous attribute $\bar{Z}$ with $dom(\bar{Z}) \in [0, 1]$. Values $z_i \in dom(Z)$, for $i=1, \ldots, c$, are encoded to values $\bar{z}_i$ in a supervised way, as the posterior probability of positives:
\begin{equation}\label{eq:perfect_encoding} \bar{z}_i = p_i \hspace{2cm} \mbox{where\ } p_i = P(Y=1|Z=z_i)
\end{equation}
However, since $P(\cdot)$ is typically unknown, an estimate of the posterior probability $p_i$ is derived from a dataset $\mathcal{D}_{tr}$ (called the \textit{training set}) of i.i.d. realizations of $Z, Y$. Let $n$ be the total number of observations, $n_i$ the number of observations where $Z=z_i$, and $n_Y$ the number of observations where $Y=1$, $n_{i, Y}$ the number of observations where $Z=z_i$ and $Y=1$. A candidate estimator consists of the observed fraction of positives among those with $Z=z_i$, hence encoding:
\begin{equation}\label{eq:encoding_estimator}
\bar{z}_i = \hat{p}_i \hspace{2cm} \mbox{where\ } \hat{p}_i = \frac{n_{i, Y}}{n_i}
\end{equation}
Such an estimator is unbiased, namely $E[\hat{p}_i] = p_i = P(Y=1|Z=z_i)$. More precisely, by Hoeffding bounds \cite{hoeffding}, for any $\epsilon > 0$, 
$P(|\hat{p}_i - p_i| \geq \epsilon) \leq  2 e^{-2 n_i \epsilon^2}$, which already points out the dependence of the estimate on the number of observations $n_i$ if $z_i$. 
Formally
, the variance of the estimator $Var[\hat{p}_i] = p_i (1-p_i) / n_i$ is relatively large when $n_i$ is small.
Unregularized target encoding does not perform well on categories with little statistical mass~\cite{catboost-encoder} as it tends to overfit the training data, failing to generalize to new data. In the extreme case of only one observation, namely $n_i = 1$, it will replace the categorical value with the target of such an observation. Such an encoding will be unrepresentative of the category and introduces a sampling (or data collection) bias at the pre-processing stage. This type of bias is what we defined as reducible bias, and can be left unnoticed because extremely small categories do not impact significantly the overall loss of the problem, but can still have an impact on fairness metrics.
%
To avoid overfitting, practitioners regularize using either $(i)$ smoothing towards the global mean, or $(ii)$ Gaussian noise, which adds normal (Gaussian) distribution noise to training data in order to decrease overfitting. Other smoothing techniques can 
 be found  in the literature, but are either 
  minimal variations of those two techniques or less  popular~\cite{category_encoders}.

\subsection{Smoothing regularization}

Smoothing towards the global mean leads to the following target encoding:
\begin{equation}
    \label{eq:re}
    \bar{z}_i = \tilde{p}_i \hspace{1cm} \mbox{where\ } 
    \tilde{p}_i = \lambda(n_i) \frac{n_{i, Y}}{n_i} + (1- \lambda(n_i)) \frac{n_Y}{n}
\end{equation}
Here, the proportion of positives among the observations with $Z=z_i$ is interpolated with the proportion of positives among all observations.
Formally, called $\hat{p}=n_Y/n$ an estimate of the prior probability $p = P(Y=1)$, we have $\tilde{p}_i = \lambda(n_i) \hat{p}_i + (1-\lambda(n_i)) \hat{p}$. The choice of the prior probability $P(Y=1)$ is natural because, lacking a sufficient number of observations for $Z = z_i$, one resorts to the proportion of positives over the whole dataset of observations.
The convex combination of the two estimators depends on $\lambda(n_i) \in [0, 1]$. The function $\lambda(\cdot)$  is assumed to be increasing with $n_i$. Intuitively, the larger the number of observations with $Z=z_i$, the more weight we give to the first estimator. Thus, the smoothed estimator is asymptotically unbiased. Conversely, the smaller the number of observations, the more weight we give to the prior probability estimator.  Thus, the smoothed estimator has a small variance for small values of $n_i$---yet, it is biased towards the prior probability.

\subsection{Gaussian noise regularization}
Gaussian noise regularization adds normal (Gaussian) distribution noise into training data after encoding the categorical attribute as in (\ref{eq:encoding_estimator}). The intuition is to perturb the data to prevent overfitting the target encoded attribute values. During the prediction stage, testing data are encoded as in (\ref{eq:encoding_estimator}) with no perturbation. Formally, called $z_{i, j}$ the $j^{th}$ occurrence of $z_i$ in the training set, $z_{i, j}$ is replaced by:
\begin{equation}\label{eq:gaussian}
\bar{z}_{i, j} = \hat{p}_{i,j} \hspace{0.6cm} \mbox{where\ } \hat{p}_{i,j} =
    \frac{n_{i, Y}}{n_i} + \epsilon_{i,j}  
    \hspace{0.6cm} \epsilon_{i,j} \sim N(0, \lambda^2)
\end{equation}
where the $\epsilon_{i,j}$'s are i.i.d.\ with mean $0$ and standard deviation $\lambda$.  Typical values for $\lambda$ are set in the range between $0.05$ and $0.6$  \cite{category_encoders}. 


\section{Theoretical Analysis}
\label{sec:theory}


We present a theoretical analysis under a number of assumptions that make it reasonably simple. First, we assume that $\bar{Z}$ is the only predictive feature. Second, we consider a probabilistic binary classifier, which for an input $\bar{Z}=\bar{z}$ outputs a score $\hat{S}(\bar{z})  \in [0, 1]$,
and a prediction $\hat{Y}(\bar{z}) = \indicator(s(\bar{z})>\nicefrac{1}{2})$. 
Third, the score is expected to approximate a Bayes optimal classifier, i.e., $\hat{S}(\bar{z}) \approx P(Y=1|\bar{Z}=\bar{z})$. For notational convenience, we write $a \bowtie b$ as a shorthand for $a > \nicefrac{1}{2} \Leftrightarrow b > \nicefrac{1}{2}$, namely $a$ and $b$ are on the same side of the decision threshold $\nicefrac{1}{2}$. We write $a \not \bowtie b$ when $a \bowtie b$ does not hold. 


\subsubsection*{\bf The case of perfect target encoding}

Under the (theoretical) assumption of knowing the true values $p_i$'s, the perfect target encoding would set $\bar{z}_i = p_i$ as in (\ref{eq:perfect_encoding}). The score $\hat{S}(\bar{z}_i) = p_i$ leads to the Bayes optimal classifier, hence maximizing AUC over the population and minimizing the classification error to the following:
%
%
\begin{equation}\label{eq:ce1}
\sum_{i=1}^c P(Z=z_i) \cdot \mathit{min}\{p_i, 1-p_i\}
\end{equation}
Consider now the equal opportunity fairness metric, namely:
\begin{equation}\label{def:eo} P(\hat{Y}=1|Y=1,\bar{Z}=\bar{z}_i) -  P(\hat{Y}=1|Y=1,\bar{Z}=\bar{z}_r)
\end{equation}
where $\bar{z}_r$ is the encoding of the reference group in the protected attribute $Z$. By definition of $\hat{Y}$, 
$\hat{Y}(\bar{z}_i)=1$ iff $\bar{z}_i = p_i > \nicefrac{1}{2}$, and analogously for $r$. Therefore, when both $p_i > \nicefrac{1}{2}$ and $p_r > \nicefrac{1}{2}$:
\[ P(\hat{Y}=1|Y=1,\bar{Z}=\bar{z}_i) =  P(\hat{Y}=1|Y=1,\bar{Z}=\bar{z}_r) = 1 \]
and then the difference is $0$. A similar conclusion is obtained when both $p_i \leq \nicefrac{1}{2}$ and $p_r \leq \nicefrac{1}{2}$. However, when the probabilities $p_i$ and $p_r$ lie on different sides of the threshold (i.e., $p_r \not \bowtie p_i$), the equal opportunity metrics is non-zero (either $-1$ or $1$). In other words, the classifier is fair only if the prediction for the reference group is the same as for the protected group. But this will impact on accuracy. In fact, assuming a constant prediction over the groups, say $Y(\bar{z}_i) = 1$, the classification error on the population becomes $\sum_{i=1}^c P(Z=z_i) \cdot (1-p_i)$, which is clearly larger than (\ref{eq:ce1}). 

In summary, even in the case of perfect target encoding and a Bayes optimal classifier, there is a  tension between error and fairness metrics optimization: the amount of unfairness is \textit{irreducible} as we assumed to know the posterior probabilities $p_i$'s, unless we admit increasing the error by not using the protected feature $Z$ in the classification problem.

\subsubsection*{\bf The case of target encoding}

Let us consider now the encoding using the (un-regularized) estimator $\hat{p}_i = n_{i, Y}/n_i$, i.e., (\ref{eq:encoding_estimator}). The score $\hat{S}(\bar{z}_i) = \hat{p}_i$ maximizes empirical AUC and minimizes the empirical error rate on the training set.
When $n_i$ is large, $\hat{p}_i \approx p_i$ (since variance of the estimator is low), and then the contribution to the classification error (\ref{eq:ce1}) and to the AUC are approximately the same as in the case of perfect target encoding. Regarding the fairness metric, we can reasonably assume that $n_r$ is large for the reference group, and then $\hat{p}_r \approx p_r$. Therefore, the equal opportunity metric is unchanged w.r.t. the case of perfect target encoding.

When $n_i$ is small, the estimate $\hat{p}_i = n_{i, Y}/n_i$ can be arbitrarily distant from $p_i$. The increment in classification error (\ref{eq:ce1}) is zero if $p_i \bowtie \hat{p}_i$, and  $P(Z=z_i) \cdot |1-2p_i|$ otherwise. Also, the AUC will be possibly smaller due to wrong ranking of instances with $Z=z_i$.
%
The equal opportunity metric is independent of $P(X=x_i)$. Compared to the perfect target encoding case, its value is unchanged if $p_i \bowtie \hat{p}_i$. Otherwise, it can either decrease (if $p_r \bowtie \hat{p}_i$) or increase (if $p_r \not \bowtie \hat{p}_i$). 

In summary, 
the variability of the estimator $\hat{p}_i$ for $n_i$ small,  negatively impacts on the performance metrics, and it propagates to the the fairness metrics, unpredictably increasing or decreasing it compared to the perfect target encoding case. The increase in the fairness metrics is  \textit{reducible} bias, which can be corrected by increasing the number of observations of $Z=z_i$.
%

\subsubsection*{\bf The case of smoothing regularization}

Let us consider now the target encoding with smoothing regularization (\ref{eq:re}). 
Let $\hat{S}()$ be the score function that minimizes the empirical error rate over the training set.
When $n_i$ is large, then $\tilde{p}_i \approx \hat{p}_i \approx p_i$, and then we fall back to the same situation as for (perfect) target encoding.

When $n_i$ is small, we have $\tilde{p}_i \approx n_Y/n \approx p$, and then instances of the training set for which $Z=z_i$  are mapped close to $\bar{Z} = p$. This does not necessarily means that the classification algorithm scores such instances as $p$ -- rather, it should score close to the mean target value of instances with $\bar{Z} = p$. Let us then be $q$ such that $\hat{S}(p) = q$. We fall back then to the reasoning for the target encoding case. The increment in classification error (\ref{eq:ce1}) is zero if $p_i \bowtie q$, and  $P(Z=z_i) \cdot |1-2p_i|$ otherwise.
Compared to the perfect target encoding case, the fairness metric value is unchanged if $p_i \bowtie q$. Otherwise, it can either decrease (if $p_r \bowtie q$) or increase (if $p_r \not \bowtie q$). 

In summary, 
the estimator $\tilde{p}_i \approx p$ for $n_i$ small is stable, but nevertheless, it can affect the performance metrics (negatively) and the fairness metrics (increase or decrease). The increase in the fairness metrics is  \textit{reducible} bias. Notice that the magnitude of the impact depends on the choice of $q$ by the machine learning algorithm under consideration, which, in principle, could be controlled for.

\subsubsection*{\bf The case of Gaussian noise regularization}

Let us now consider the Gaussian noise regularization (\ref{eq:gaussian}).  
Its expectation is $E[\hat{p}_{i,j}] = E[\hat{p}_i] + E[\epsilon_{i, j}] = p_i$, hence the estimator is unbiased. Its variance is $Var[\hat{p}_{i,j}] = Var[\hat{p}_i] + \lambda^2$. From this, we have that: (1) the variance is larger than in the case of target encoding, and, a fortiori, of the smoothing regularization; (2) the larger the regularization parameter $\lambda$, the larger the variance.  
Let us consider a partition of the instances with $Z=z_i$ based on whether $\hat{p}_{i,j} \bowtie p_i$ holds or not. 

For the subset $\hat{p}_{i,j} \bowtie p_i$, there is no change in classification error, nor in the equal opportunity fairness metrics, when compared to the perfect target encoding case. 

Consider instead the subset $\hat{p}_{i,j} \not \bowtie p_i$. 
The increment in classification error (\ref{eq:ce1}) is  $\sum_{\bar{z}} P(Z=z_i, \bar{Z}=\bar{z}, \bar{z} \not \bowtie p_i) \cdot |1-2p_i|$. For $n_i$ small, this is lower than in the cases of target encoding and smoothing regularization. For $n_i$ large, this is greater than in those two cases, where it is $\approx 0$. However, since $Var[\hat{p}_i] \approx 0$, this case only occurs for a large $\lambda^2$ that causes crossing the decision boundary, i.e., for which $\hat{p}_{i,j} \not \bowtie p_i$. 
Compared to the perfect target encoding case, the fairness metric can either decrease (if $p_r \bowtie \hat{p}_{i,j}$) or increase (if $p_r \not \bowtie \hat{p}_{i,j}$). Again, for small $n_i$'s the impact is smaller than for target encoding and smoothing regularization, and  for large $n_i$'s, this can only occur if $\lambda^2$ is large enough for crossing the decision boundary.

In summary, Gaussian noise regularization adds some controllable variability that impacts mainly on small $n_i$'s and for a subset of the data distribution for which a random perturbation may cross the decision boundary. If this happen, there is an increase in classification error, and some chance to increase/decrease the equal opportunity fairness metric. The increase in the fairness metrics is  \textit{reducible} bias.

\subsubsection*{\bf The case of one-hot encoding}

Consider a variant of one-hot encoding setting $\bar{z}_i=2^i$, i.e., mapping $z_i$ into a number with the $i$-{th} digit set to $1$ and all others to $0$. Such a variant keeps our assumption of one predictive feature only. The previous sub-sections on perfect target encoding and on target encoding could be repeated, almost unchanged, as they only require $\hat{S}(\bar{z}_i) = p_i$ and $\hat{S}(\bar{z}_i) = \hat{p}_i$ respectively, ignoring the form of the coding of $\bar{z}_i$. We would therefore expect that the behavior of one-hot encoding and (unregularized) target encoding be very similar. What can make a difference is that most machine learning algorithms treat one-hot encoding as a collection of i.i.d. features, ignoring their dependencies (i.e., that one and only one digit must be $1$). This may lead to an greater classification error when compared to target encoding.

\section{Experiments} 
\label{sec:experiments}

In this section, we study the implications of model accuracy and fairness when encoding categorical protected attributes. \emph{(H1)} The first main hypothesis is that encoding the protected attribute helps to improve accuracy. \emph{(H2)} The second main hypothesis is that fairness is worsened by encoding.  To evaluate both (H1) and (H2) we compare two encoding methods one-hot encoding and target encoding versus not encoding the protected attribute. Our third hypothesis \emph{(H3)} is that target encoding regularization can improve fairness without significantly impacting predictive performance, and we evaluate this by comparing two regularization techniques across various hyperparameters as part of the machine learning pipeline's preprocessing step. Additionally, in the last section, we explore the effects of intersectional protected categorical attributes, which augment the previous three hypotheses.\emph{(H3)}


\subsection{Experimental Setup}
\subsubsection{Datasets: COMPAS and FolkTables.}
\label{sec:experiments.compass}
We have chosen two datasets that happen to exhibit high-cardinality sensitive categorical attributes in a binary classification problem: COMPAS~\cite{compass} and FolkTables~\cite{ding2021retiring}. We report our method and findings on the COMPAS dataset in the main body of this paper and apply the same methodology on FolkTables, but report findings from the latter in the appendix. Overall, the findings are very similar in both datasets.

COMPAS is an acronym for Correctional Offender Management Profiling for Alternative Sanctions, which is an assistive software and support tool used to predict the risk that a criminal defendant will re-offend. The dataset provides a category-based evaluation labeled as high risk of recidivism, medium risk of recidivism, or low risk of recidivism. We convert this multi-class classification problem into binary classification by combining the medium risk and high risk of recidivism and comparing them to low risk of recidivism. The input used for the prediction of recidivism consists of 11 categorical attributes, including gender, custody status, legal status, assessment reason, agency, language, ethnicity, and marital status. The sensitive attribute that we consider is \textit{Ethnic} for the single discrimination case, whose protected group we define as the most represented group: African-American (cf. Figure~\ref{fig:pieCompas}).

To study fairness related to intersectional attributes, we created the variable \textit{EthnicMarital}, engineered by concatenating \textit{Ethnic} and \textit{Marital} status. This new attribute has a high cardinality of 46 distinct values (cf. Figure~\ref{fig:pieCompas}). The most predominant category is \textit{African-American Single}, and it will be the protected group (cf. Figure~\ref{fig:pieCompas}) for the intersectional fairness case. To compare disparate treatment between groups we will make use of  \textit{Caucasian Married} as the reference group. It is worth noting that the contribution of the attributes to the model performance, based on attribute importance explanation mechanism\cite{shapTree,ribeiro2016why,christoph_molnar2019_interpretableML,desiderataECB}, is highly relevant. The available data is split into a 50/50 stratified train/test split, maintaining the ratio of each category between train and test set.
In the Figure~\ref{fig:pieCompas} of the appendix, we can see how the group distributions are unbalanced  with two groups, \textit{African-American} and \textit{Caucasian}, that account for the $+80\%$ of the data. For the intersectional fairness case, the number of groups increases, making room for more distinct, disparate, and imbalanced groups~\cite{DBLP:conf/sdm/FouldsIKP20}.
  
\subsubsection{Machine learning algorithms} Our experiments involve  a logistic regression model, a neural network (Multi-layer Perceptron classifier), and a gradient-boosting decision tree. All models are trained on the training set. These three models provide examples of a model with large bias (the linear regression model), a highly complex model (the MLP classifier), and the extensively used, state-of-the-art  gradient-boosting decision tree~\cite{supervised_learning_algorithms,gbm_travel,gbm,gbm_nn,get_bristol}.

\subsubsection{Choice of metrics and models}
\label{sec:methodology.metrics}
\textbf{\\ Model performance metrics.} Previous work on fair machine learning have evaluated their experiments on COMPAS using accuracy as a performance metric~\cite{zafat2017fairnessDisparate,zafar2017parity,zafar2017fairness}, but given that we want to study effects of group imbalance, we consider accuracy to be a less informative measure of model performance. 

Area Under the Curve (AUC) measures the diagnostic ability of a binary classifier as its discrimination threshold is varied. AUC is less susceptible to class imbalance than accuracy or precision and also accepts soft probabilities predictions~\cite{statisticallearning}. 
An AUC of 0.5 is equal to random predictions.

\textbf{Fairness metrics} We use three different metrics $\ell_{i,j}(f,X,y)$ to judge fairness of classifier $f$ on data $X$ between groups indexed by $i,r$ and we denote $\hat{Y} = f(X)$ for simplicity:

\begin{itemize}
    \item \textbf{Statistical Parity (Strong Demographic Parity)}: The difference between favorable outcomes received by the unprivileged group and privileged group ~\cite{kamishima2012fairness,DBLP:conf/icml/ZemelWSPD13,DBLP:conf/kdd/FeldmanFMSV15,DBLP:conf/kdd/Corbett-DaviesP17}. DP ensures that a fair decision does not depend on the sensitive attribute regardless of the classification threshold used ~\cite{DBLP:conf/uai/JiangPSJC19,DBLP:conf/aaai/ChiappaJSPJA20} 
    \begin{equation}
    \text{DP}_{i,j} = d(P(\hat{Y}|Z=i),P(\hat{Y}|Z=r))
    \end{equation}
    
    where $d(\cdot,\cdot)$ is a distance function. In this work, we use the Wasserstein distance as a measure between the two probabilistic distributions. The intuition behind Demographic Parity is that it states that the proportion of each segment of a protected attribute should receive a positive outcome at equal rates, a positive outcome is a preferred decision.
    
    \item  \textbf{Equal opportunity fairness.} Following~\citet{DBLP:conf/nips/HardtPNS16}'s emphasis on ensuring fair opportunity instead of raw outcomes, we choose \emph{equal opportunity} (EO) as a fairness notion and use the metric {\it disparate treatment} (difference between the true positive rates) to measure unfairness, which is estimated using the disparate treatment metric~\cite{zafat2017fairnessDisparate}. For simplicity, we refer to the interplay of these concepts as the \textit{equal opportunity fairness (EOF)} metric. The value is the difference in the True Positive Rate (TPR) between the protected group and the reference group~\cite{bigdata_potus,bigdata_potus1}). 
    
    \begin{gather}\label{eq:TPR}
    \text{TPR}_i = P(\hat{Y}=1|Y=1,Z=i)\quad
    \text{EOF}_{i,r}= \text{TPR}_i - \text{TPR}_r
    \end{gather}
    
    A negative value in (\ref{eq:TPR}) is due to the worse ability of a Machine Learning model to find actual recidivists for the protected group (i) in comparison with the reference group (j). 

    \item \textbf{Average Absolute Odds (Equalized Odds)}: The sum of the absolute differences between the True Positive Rates and the False Positive Rates of the unprivileged group plus the same ratio for the  privileged group.
    
    \begin{gather}\label{eq:AAO}
    \text{FPR}_{i} = P(\hat{Y}=1|Y=0,Z=i)\\
    \text{AAO}_{i,r}
    = \frac{1}{2}(|FPR_i - FPR_r| + |TPR_i - TPR_r|)
    \end{gather}
    The  intuition is that an $AAO=0$ means the algorithm is fair because it results in the same False Positive Rate and True Positive Rate for the privileged group as an unprivileged group. If the algorithm causes a difference in either, then $AAO\neq0$. A deviation in each term contributes equally to AAO, then False Positives Rates might have different social implications than True Positives Rates~\cite{2018aequitas,DBLP:journals/csur/MehrabiMSLG21,wachter2020bias}.
\end{itemize}

All three metrics indicate better fairness between groups $i,j$ by values closer to 0.
We calculate the overall fairness $\mathcal{L}$ of the model $f$  
on data of interest $X$ given a fairness metrics $\ell$, reference group $i$ and other  groups $\{i | i\neq r\}$ as:

\begin{equation}
    \mathcal{L}(f,X,y,i,r) = \sum_{i \neq r} |(\ell_{i,r}(f,X,y)|
\end{equation}

where each group $i$ contributes equally to the overall metric, meaning these are not weighted by the number of individuals in each group. 


\subsection{Experimental results: encoding categorical protected attributes}
\label{exp:first}



In this section we evaluate hypotheses (H1), (H2), and (H3). The trade-offs between fairness metrics and predictive performance metrics (AUC) are analyzied using two different encoding techniques (Section~\ref{sec:relatedwork}), with two different regularization techniques (Section~\ref{sec:methodology}) and two different estimators (Section~\ref{sec:methodology.metrics}).  The ranges of the regularization hyperparameters are:  $\lambda \in$ $[0,5]$ for the width of the Gaussian noise regularization;  $m \in [0,1000000]$ for the additive smoothing using the $m$-probability estimate function $\lambda(n_i) = n_i/(n_i+m)$ (see~\cite{high-cardinality-categorical}). These hyperparameters will also be kept for the rest of the experiments for this dataset. 

\begin{figure*}[ht]
    \centering
    \includegraphics[width=1\linewidth]{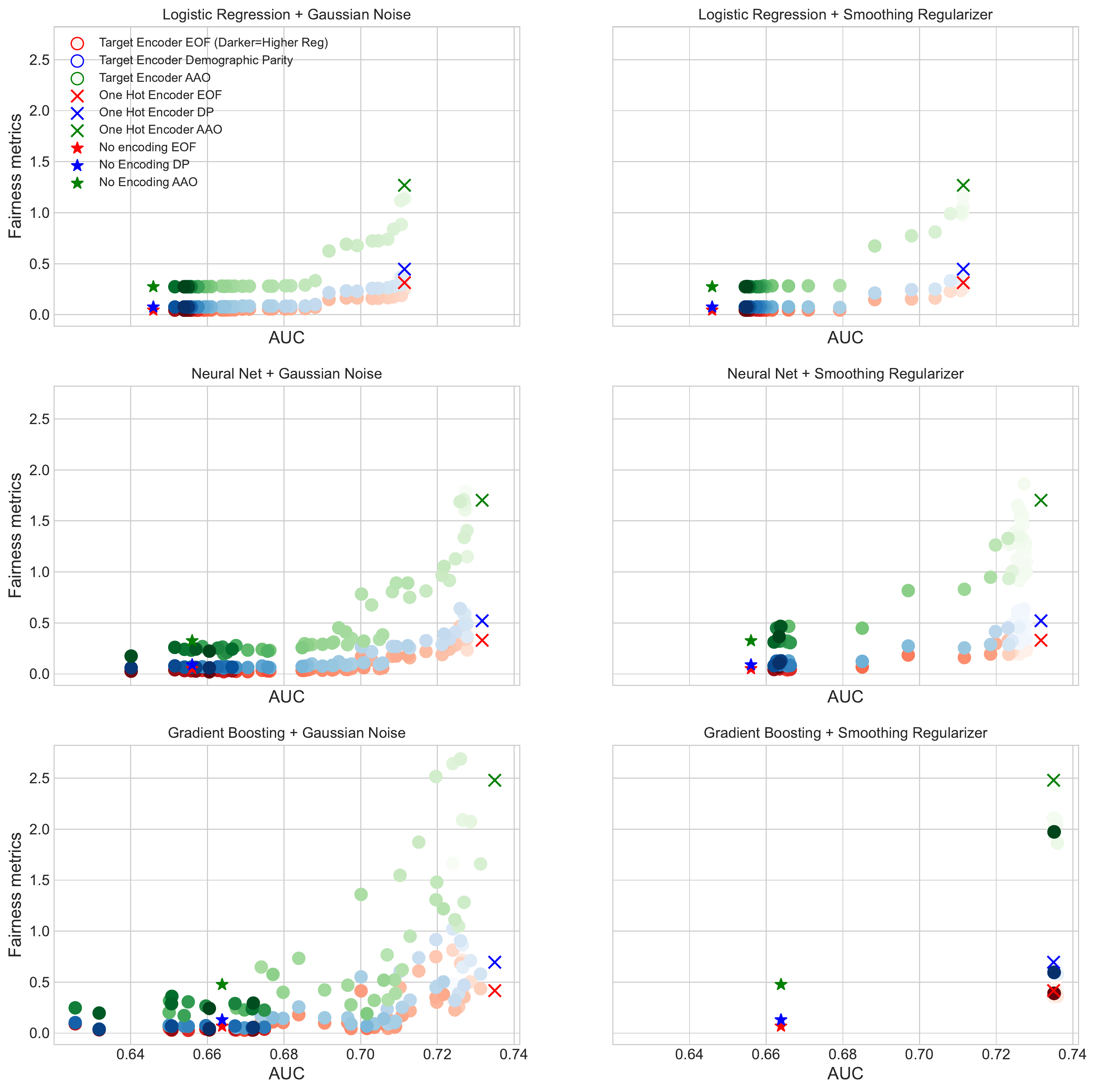}
    \caption{Comparing one-hot encoding and target encoding regularization (Gaussian noise and smoothing) for the Logistic Regression,Neural Network, and Gradient Boosting classifiers over the test set of the COMPAS dataset. Colored dots regard different regularization parameters: the darker the red the higher the regularization. Different colors imply different fairness metrics. Crossed dots regard one-hot encoding and starred dots are the results of models that exclude the use of the protected attribute.}
    \label{fig:compas1}
\end{figure*}

\begin{figure}[ht]
    \centering
    \includegraphics[width=1\linewidth]{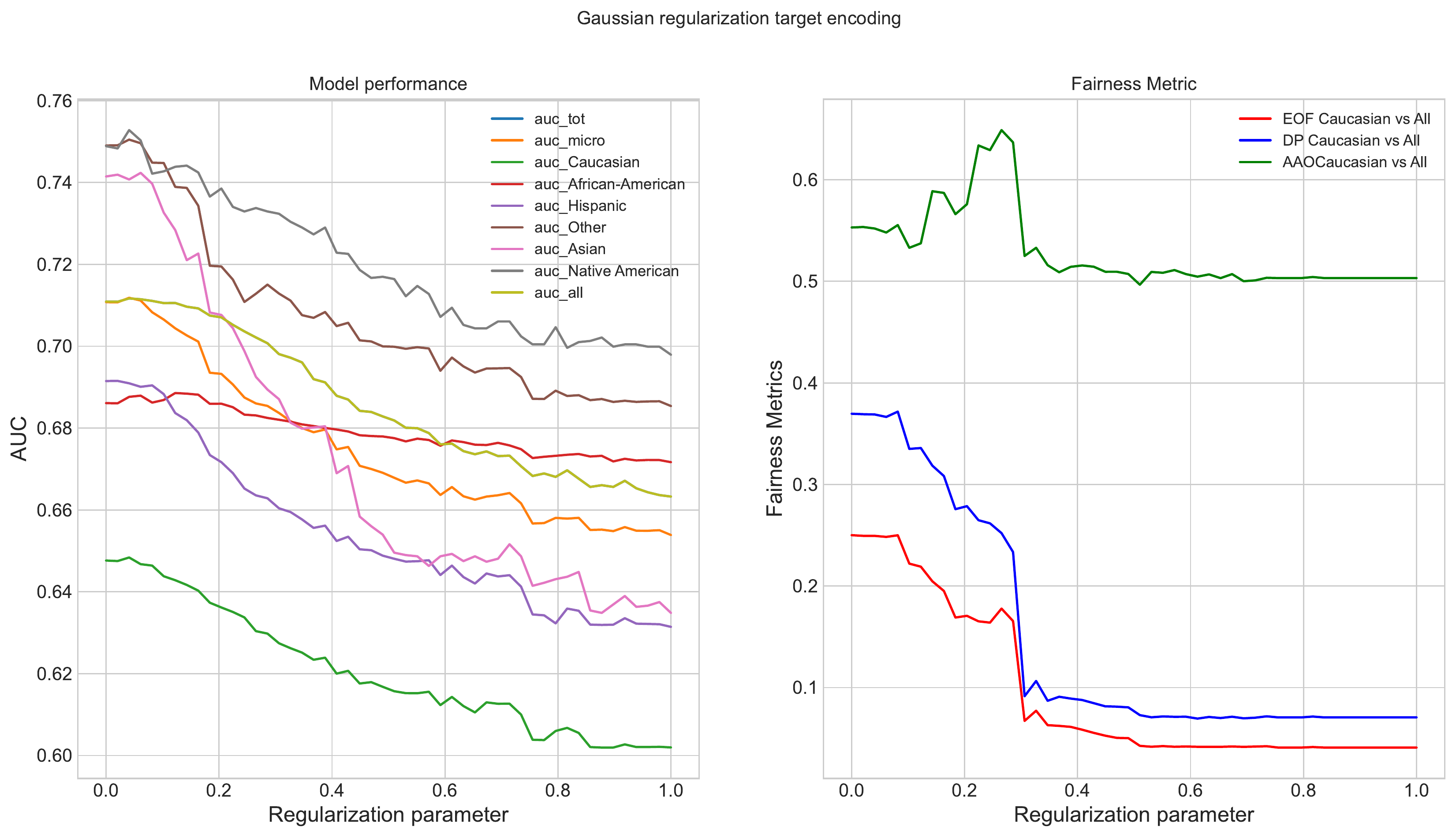}
    \caption{Impact of the Gaussian noise regularization parameter $\lambda$ on performance and fairness metrics over the test set of the COMPAS dataset using a Logistic Regression with L1 penalty. In the left image the AUC of the all the protected groups over the regularization hyperparameter. On the right, the equal opportunity fairness, demograpic parity and average absolute oods variation throughout the regularization hyperparameter.
    }
    \label{fig:hyperCompas}
\end{figure}

Under \textbf{Gaussian noise regularization} (cf Figure \ref{fig:compas1} left images), evaluation support our three hypotheses: (H1)  predictive performance improves when encoding the categorical protected attributes. In all six experiments the improvements reported are in the range of [0.04-0.8] AUC. (H2)   all the experiments exhibit  fairness degradation, up to one order of magnituce. (H3) We  observe that within low regularization ranges of hyperparameters (lighter dots),  fairness improves without compromising the predictive performance of the model. However, for higher levels of regularization (darker dots), fairness metrics have a plateau while predictive performance (AUC) keeps degrading. At the highest regularization penalty, target encoding often matches performance and fairness with \enquote{no encoding} while with no regularization  matches \enquote{one hot encoding}.  Later in this section, we discuss this in depth.

We find similar results in the case of \textbf{smoothing regularization} (cf Figure \ref{fig:compas1} right images). But not for our regularization hypothesis. While it should for the linear regression and the neural networks, it does not for the gradient-boosting decision trees, whose target encoding regularization effects are negligible in both fairness and model performance. These can be due to smoothing producing a shrinking effect where decision tree-based models are generally not affected by monotonous attribute transformations \cite{xgboost}.

In Figure \ref{fig:hyperCompas}, we analyze deeper the target encoding hyperparameter fairness-accuracy trade-off. We can see that there is an optimal trade-off value around $0.3$, where the equal opportunity fairness and demographic parity have dropped down toward the fairness plateau and the model performance has only slightly decreased. The predictive performances (AUC) of different groups have different negative slopes, ethnic groups as \textit{Asian} or \textit{Native-American} have a drastic drop in performance while groups as \textit{African-American} have only a small performance decay. \textit{African-American} represents the $44.4\%$ of the data while \textit{Asian} or \textit{Native-American} do not achieve even a statistical representation of $1\%$. 


Respect to the accross model comparison, the predictive performance of the gradient boosting decision tree model is best, followed by the neural network and then the linear model \cite{grinsztajn:hal-03723551,BorisovNNtabular}.
From the fairness perspective, more complex models have a stronger fairness violation.


\subsection{Experimental results: Engineering intersectional fairness}
\label{exp:intersectional}

\begin{figure}[ht]
    \centering
    \includegraphics[width=1\linewidth]{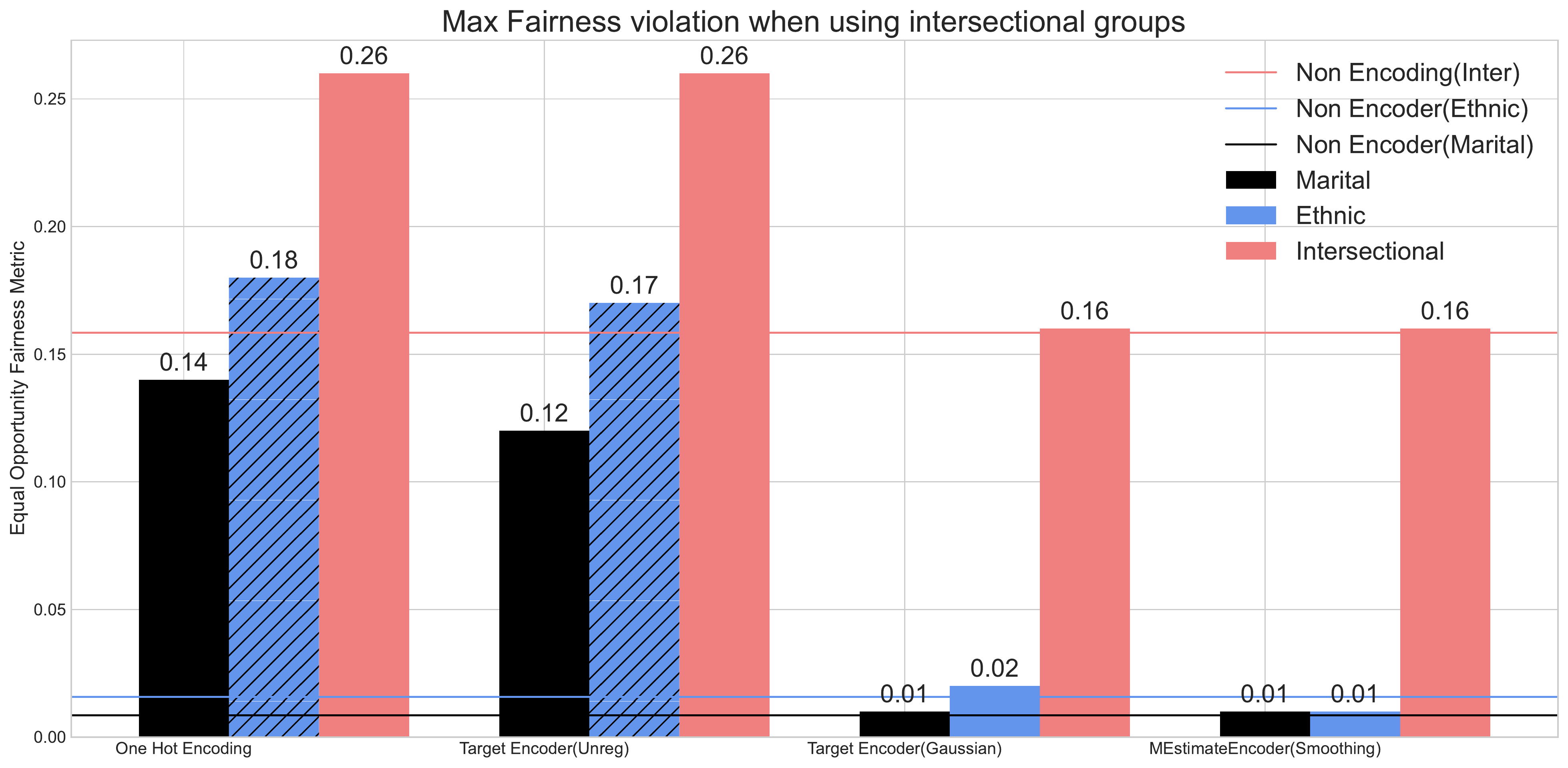}
    \caption{Equal opportunity fairness implications of encoding categorical protected attributes and their regularization effects on the Compas Dataset. Horizontal lines are the base lines where the protected attribute is not included in the training data. Regularized target encoding does not harm fairness metrics but it can improve predictive performance on this dataset.}\label{fig:InterCompas}
\end{figure}

Our intersectional fairness hypothesis are that $(i)$ the engineering of intersectional features degrades fairness, $(ii)$ that encoding the categorical protected attribute increases discrimination and $(iii)$
that by regularizing target encoding we can reduce intersectional discrimination to no-encoding levesl.

To provide evidence of the potential effects of encodings on intersectional fairness, we concatenate \textit{Ethnic} and \textit{Marital} status of the COMPAS dataset.  We select \textit{Caucasian Married} as the reference group and compare what is the maximum fairness violation between all groups. For visualization purposes, we select the generalized linear model or previous section and we focus on the notion of Equal Opportunity Fairness since we have seen in the previous experimental section that the three fairness metrics exhibit the same behaviour. 

In Figure \ref{fig:InterCompas}, we can see how attribute concatenation creates intersectional attributes and boosts fairness violations. Validating our first hypothesis that fairness metrics increase just by the engineering of intersectional discrimination. Even in when there is no-encoding the protected attributes (horizontal lines), the maximum fairness violation between groups is increased by an order of magnitude from $0.015$ for \textit{Ethnic} or $0.08$ for \textit{Marital Status} to $0.16$ for the intersectional attribute of both. The increase of discrimination when engineering intersectional protected attributes aligns with the social findings presented originally back in $1958$ when Kimberle Crenshaw ~\cite{crenshaw2013demarginalizing} wrote her critique to the anti-discrimination doctrine, feminist theory and anti-racist politics, to describe the ways in which different forms of oppression intersect and compound one another, increased discrimination for marginalized groups.

Our second hypothesis is also validated as both encoding techniques achieve a higher equal opportunity violation than no-encoding the protected attribute. Finally we can see that by regularizing the target enconding of protected attribute fairness can be imporved.  This is not surprising, and, in general, attribute concatenation can worsen fairness both on the side of irreducible bias (because $p_i$ and $p_r$ become more distant) and on the side of reducible bias (because $n_i$ becomes smaller), as we have seen in the theoretical section.

\section{Conclusion}
\label{sec:conclusion}

In this work, we have focused on how the encoding of categorical attributes can reconcile model quality and fairness. We have provided theoretical and empirical evidence that encoding categorical attributes could induce two different types of bias: an \textit{irreducible bias}, due to the learning of discriminant information between the protected and reference groups, and a \textit{reducible bias} due to the large variance sampling found in small protected groups. 

Through theory and experiments, we showed that the most used categorical encoding method in the fair machine learning literature, one-hot encoding, consistently discriminates more than target encoding. However, we found some promising results using target encoding. Target encoding regularization showed fairness improvements with the risk of a noticeable loss of model performance in the case of over-parametrization. We also found that the type of regularization chosen is relevant depending on the algorithm used. These results support our view that (regularized) target encoding can be a useful tool for fair machine learning. Furthermore, we discussed how attribute engineering can boost the performance of machine learning algorithms but can lead to fairness violations increase, potentially due to both reducible and irreducible biases.

These experiment aims to motivate industry practitioners, where in many situations the usage of the protected attribute is not strictly prohibited, that with slight changes on the encoding of the protected attribute improvements on fairness can be achieved without any noticeable detriment of predictive performance

\textbf{Limitations and disclaimer:} In this work we have used two models, two encodings, two regularization techniques, and two datasets. In order to do a large-scale comparison we need to choose a single scalar metric that accounts for the trade-off between model accuracy and model fairness. Also, encodings are more impactful the more the protected attribute is related to the target variable. This work aims to show what are some of the implications of encoding categorical protected attributes at any moment it should be understood as if in any situation encoding categorical protected attributes won't increase fairness metrics, we advocate that considering the effects of encoding regularization along the fairness axis too, not only on the predictive performance axis. The usage of fair-AI methods does not necessarily guarantee the fairness of AI-based complex socio-technical systems~\cite{DBLP:conf/fat/KulynychOTG20,DBLP:conf/fat/ScottWMDSB22}.

\subsection*{Reproducibility Statement}
We make our results are open-source and reproducible: original data, data preparation routines, code repositories, and methods are all publicly available at \url{https://anonymous.4open.science/r/FairEncoding-72B0/README.md}. Note that throughout our work, we do not perform any hyperparameter tuning (except on the regularization) instead, we use default \texttt{scikit-learn}
hyperparameters~\cite{scikit-learn}. We describe the system requirements and software dependencies of our experiments. Our experiments were run on a 4 vCPU's server of 15GB of RAM.

\section*{Acknowledgments}
This work has received funding by the European Union’s Horizon 2020 research and innovation programme under the Marie
Skłodowska-Curie Actions (grant agreement number 860630) for
the project : \enquote{NoBIAS - Artificial Intelligence without Bias}. Furthermore, this work reflects only the authors’ view and the European Research Executive Agency (REA) is not responsible for any
use that may be made of the information it contains. 
The authors would like to thank Gourab K. Patro for its early stage contributions.

\bibliographystyle{apalike}
\bibliography{ref}
\newpage
\section*{Appendix: Experiment results}\label{sec:appendix}
\subsection*{Data: Compas data overview}

\begin{figure}[ht]
\centering
\includegraphics[width=1\linewidth]{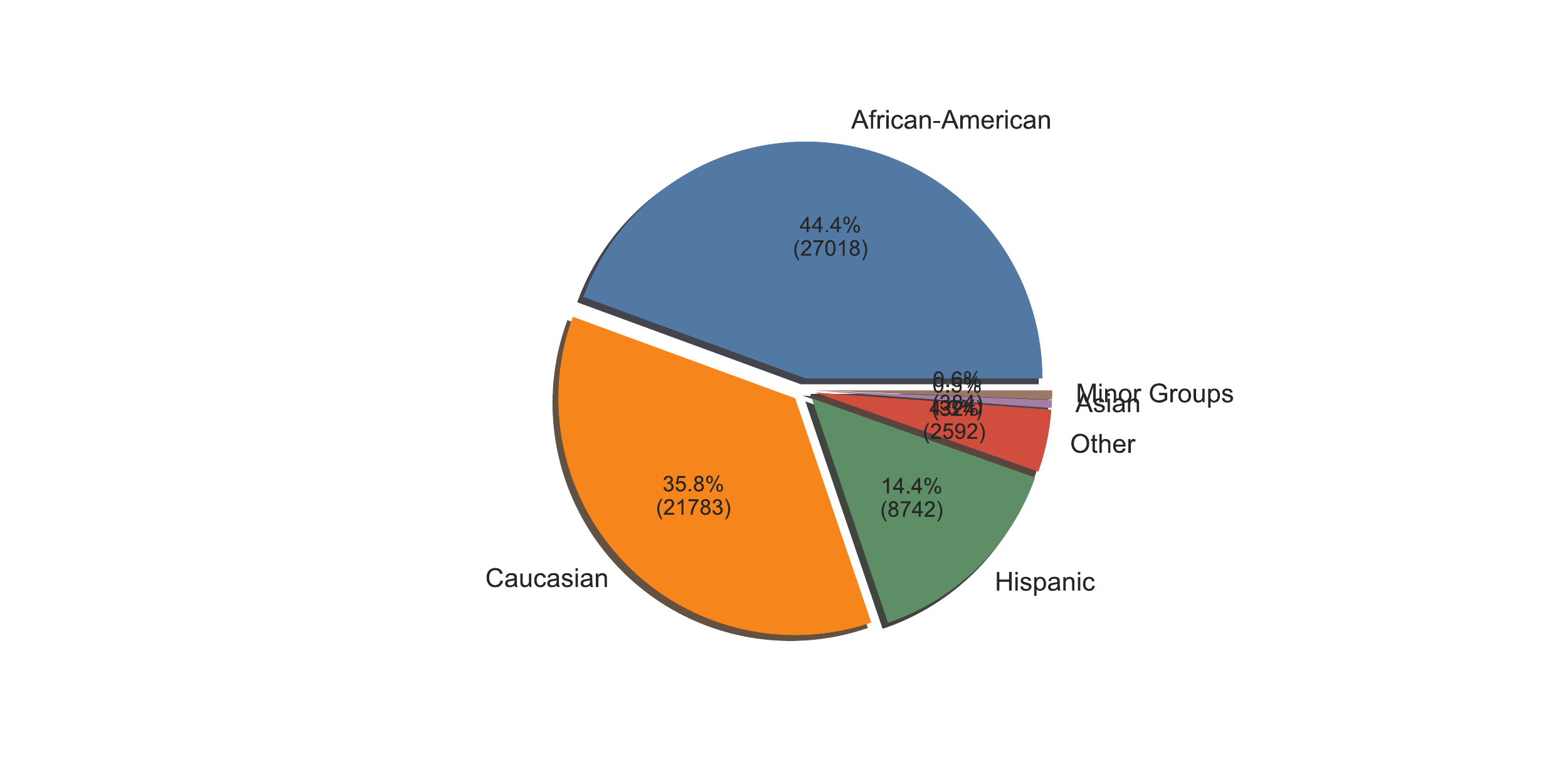}
\caption{Distribution of the protected attribute categories to be encoded and regularized for the COMPAS data~\cite{compass}. Predominant \textit{Ethnic} categories are \textit{African-American} and \textit{Caucasian}}\label{fig:pieCompas}
\end{figure}

In Figure~\ref{fig:pieCompas} we compliment the experimental section on the main body of the paper by showing the distributions of the ethnic groups. There are two groups (\textit{African-American} and \textit{Caucasian}) that account for the $80\%$ of the data, while there are less represented groups such as Asian or Arabic that have a less significant statistical weight. For the intersectional fairness case, the number of groups is increased to 46 distinct groups, making room for more distinct, disparate, and imbalanced groups\cite{DBLP:conf/sdm/FouldsIKP20}. 

\begin{figure}[ht]
\centering
\includegraphics[width=1\linewidth]{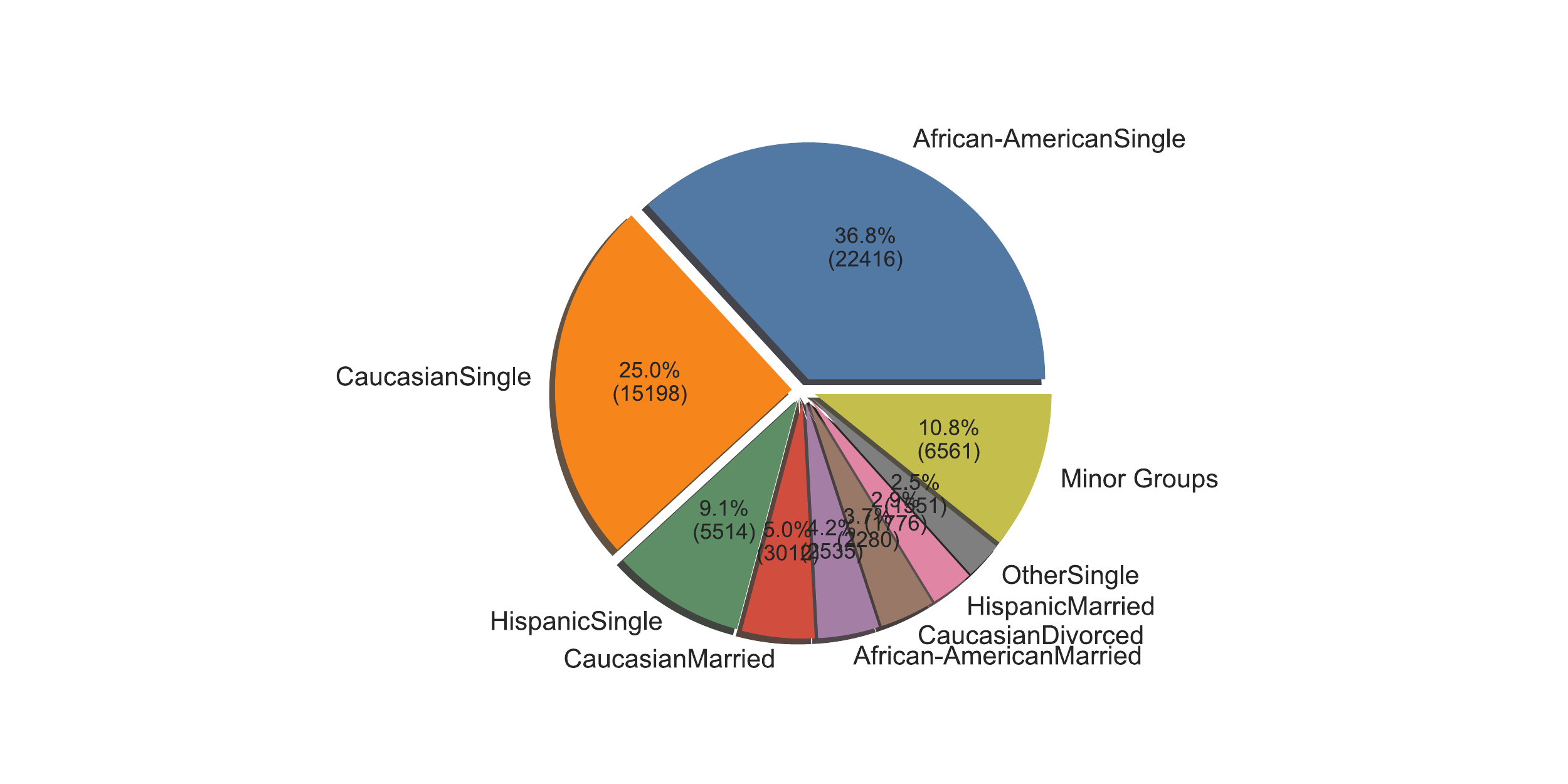}
 \caption{
 Distribution of the intersectional protected attribute \textit{Ethnic-Marital} to be encoded and regularized for the COMPAS data~\cite{compass}. Predominant categories is  categories distribution are \textit{African-American Single} and \textit{Caucassian Single}}
\end{figure}

\subsection*{Data: US Census Income}\label{data:folks}

In this section, we provide experiments on the Adult Income data set\footnote{Please see the ACS PUMS data dictionary for the full list of variables available \url{https://www.census.gov/programs-surveys/acs/microdata/documentation.html}} derived from the US census data~\cite{ding2021retiring}. Folktables package provides access to data-sets derived from the US Census, facilitating the bench-marking of fair machine learning algorithms. We select the data from  California in 2014 that covers 60,729 individuals including their race, that has 8 unique groups. Aiming to predict predict whether an individual's income is above $50,000$. The data is split into a 50/50 train/test split, maintaining the ratio of each category between train and test set.

\begin{table}[ht]
\begin{tabular}{l|ll}
                & Distribution & Ratio($\%$) \\ \hline
White           & 117209       & 0.66  \\
Asian           & 28817        & 0.16  \\
Other           & 20706        & 0.11  \\
Black           & 8435         & 0.05  \\
Native          & 1121         & 0.005 \\
Hawaiian        & 612          & 0.003 \\
American Indian & 379          & 0.002
\end{tabular}
\caption{Statistical distribution of the protected attribute \textit{Race} on the US census dataset.}
\end{table}

\begin{figure}[ht]
    \centering
    \includegraphics[width=1\linewidth]{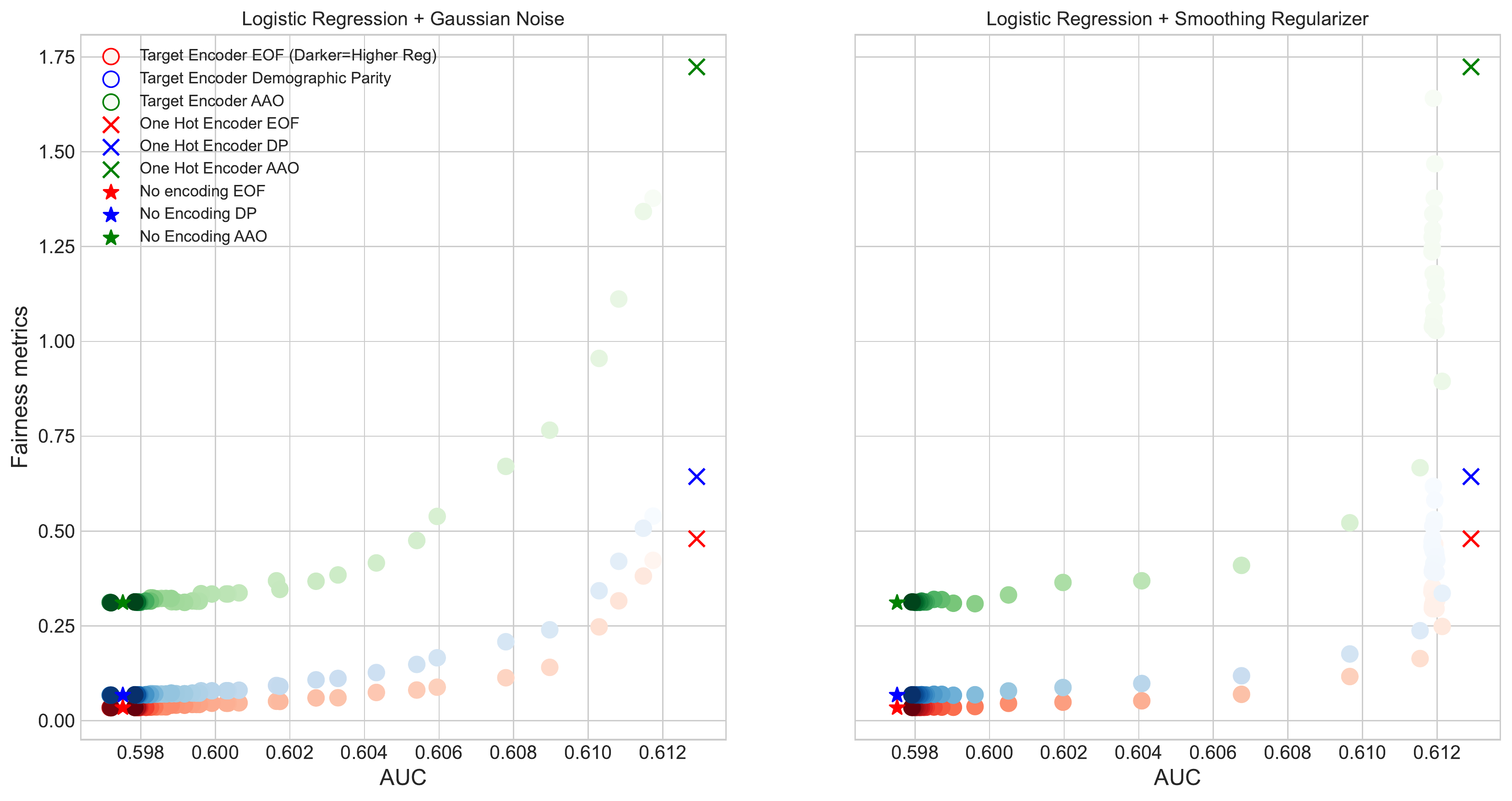}
    \caption{Comparing one-hot encoding and target encoding regularization (Gaussian noise and smoothing) for the Logistic Regression classifier over the test set of the US Income dataset. The Reference group is \textit{White}. Colored dots regard different regularization parameters: the darker the red the higher the regularization. Different colors imply different fairness metrics. Crossed dots regards one-hot encoding and starred dots not including the protected attribute in the data.}
    \label{fig:folks1}
\end{figure}

\begin{figure}[ht]
    \centering
    \includegraphics[width=1\linewidth]{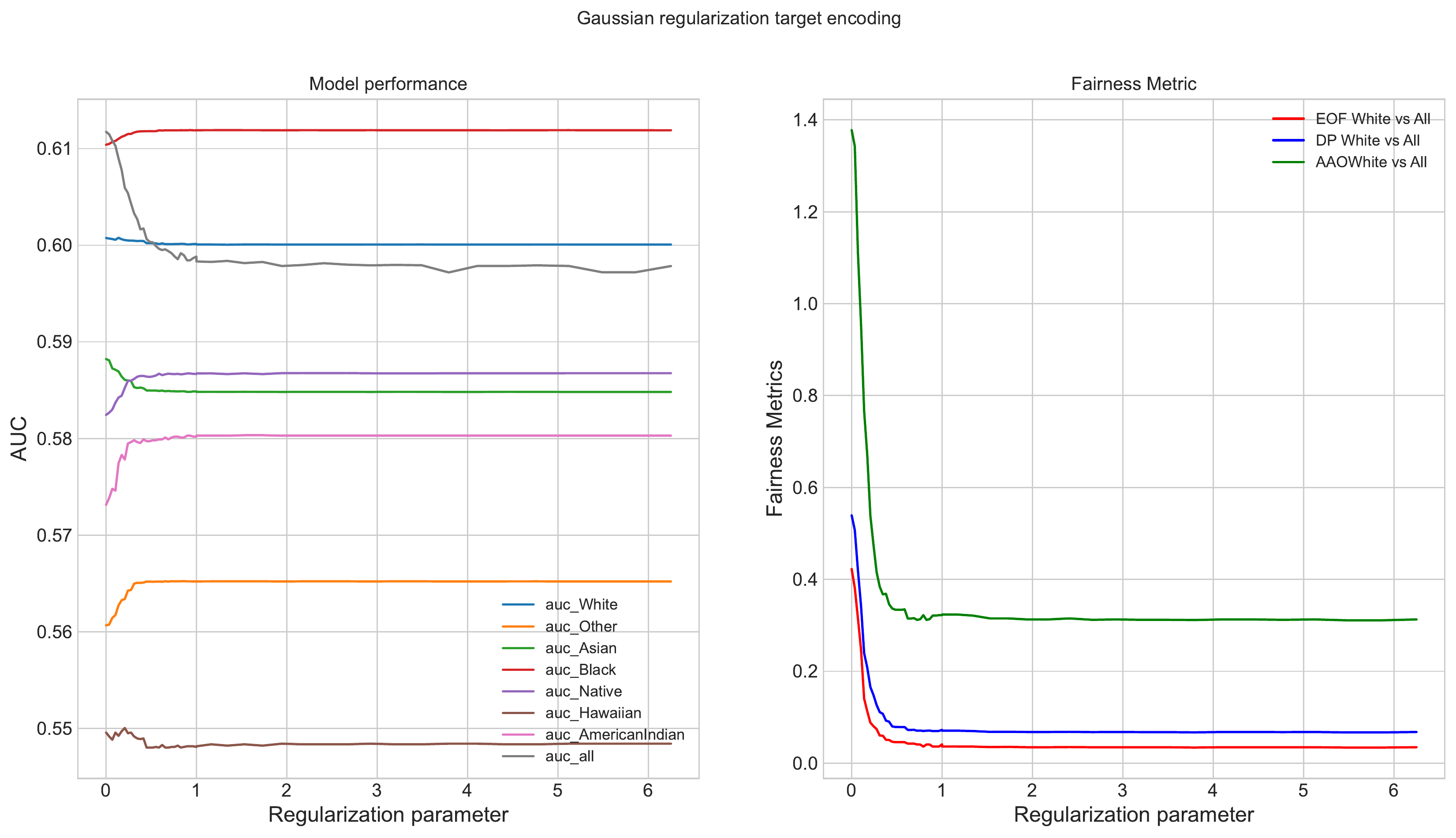}
    \caption{Impact of the Gaussian noise regularization parameter $\lambda$ on performance and fairness metrics over the test set of the US income dataset using a Logistic Regression with L1 penalty. In the left image the AUC of the all the protected groups over the regularization hyperparameter. On the right, the equal opportunity fairness, demograpic parity and average absolute oods variation throughout the regularization hyperparameter.
    }
    \label{fig:hyperfolks}
\end{figure}

Under \textbf{Gaussian noise regularization}(left images of Figure \ref{fig:folks1}), for the logistic regression we can validate our three hypothesis: (H1) first that predictive performance improves when encoding the categorical protected attributes, in this case respect to the results on Compass dataset, the AUC improvements are smaller, this can be due to the lack of predictive power of the categorical protected attribute. (H2) that fairness metrics are worsened by the encoding of the protected attribute, the differences between no-encoding versus one-hot encoding or non-regularized target encoding are substantial. 

Our last hypothesis, (H3) is that through regularizaton predictive performance can be improved without compromising the fairness of the model. We can observe that during low regularization range of hyperparameters (lighter dots),there are high fairness violation with only a small improvement on predictive performance. On the other side, for high regularization (darker dots), fairness metrics have a smaller value. At the highest regularization penalty, target encoding often matches performance and fairness with \enquote{no encoding} while with no regularization  matches \enquote{one hot encoding}.

\end{document}